# ChatGPT, Let us Chat Sign Language: Experiments, Architectural Elements, Challenges and Research Directions

Nada Shahin
Intelligent Distributed Computing and Systems (INDUCE) Lab,
Department of CS and Software Engineering, College of IT, National Water and Energy Center,
UAE University, Al-Ain, UAE
700039075@uaeu.ac.ae

Leila Ismail
Intelligent Distributed Computing and Systems (INDUCE) Lab,
Department of CS and Software Engineering, College of IT, National Water and Energy Center,
UAE University, Al-Ain, UAE
CLOUDS lab, School of Computing and Information Systems,
The University of Melbourne, Melbourne, Australia
leila@uaeu.ac.ae

*Abstract*— *ChatGPT is a language model based on Generative AI. Existing research work on ChatGPT focused on its use in various domains. However, its potential for Sign Language Translation (SLT) is yet to be explored. This paper addresses this void. Therefore, we present GPT's evolution aiming a retrospective analysis of the improvements to its architecture for SLT. We explore ChatGPT's capabilities in translating different sign languages in paving the way to better accessibility for deaf and hard-of-hearing community. Our experimental results indicate that ChatGPT can accurately translate from English to American (ASL), Australian (AUSLAN), and British (BSL) sign languages and from Arabic Sign Language (ArSL) to English with only one prompt iteration. However, the model failed to translate from Arabic to ArSL and ASL, AUSLAN, and BSL to Arabic. Consequently, we present challenges and derive insights for future research directions.*

## I. INTRODUCTION

Deaf and hard of hearing (DHH) constitute a significant percentage of the global population. According to the World Health Organization (WHO), 430 million individuals have hearing loss worldwide [1]. However, there are only around 30,000 employed interpreters in the United States [2], which is very low compared to the number of DHHs within the same country (37.5 million [3]). Therefore, sign language interpretation is critical. While Natural Langue Processing (NLP) has helped to transform spoken and written languages, there have been research efforts to expand its use to Sign Language Translations (SLT) [4] by converting a text to SL and vice-versa. However, due to the sophisticated structure of sign language gestures and the complexity of its grammar, SLT poses unique challenges [5]. In this paper, we test ChatGPT's (Chat Generative Pre-Trained Transformer) ability to develop SLT applications while considering the grammar and syntax constraints of sign language.

ChatGPT is an artificial intelligent generated content (AIGC) developed by OpenAI that produces text as a response based on its input, whether it is audio, text, or images. The model combines GPT, a language model based on a decoder transformer, and reinforcement learning using the Proximal Policy Optimization (PPO) algorithm and human feedback [6]. It has the characteristic of keeping the history of what it learned during a conversation and using it along with the current input to produce a context-aware answer. This is thanks to its attention layers as part of its architecture [7]. Furthermore, ChatGPT was pre-trained on a massive text corpus, including books, articles, websites, and conversational data from online forums and chat rooms [8]. This enabled the model to grasp the patterns and structure of natural language. It can be used for natural language processing tasks such as language translation and summarizing, various conversational applications including question answering, and performing sentiment analysis (positive, negative, and neutral) [8]. ChatGPT can also generate, correct, and help with programming coding based on its understanding of programming concepts. Such a task is completed based on the pre-training of programming languages and frameworks. Despite its potential, in some cases, ChatGPT produces incorrect results. This is critical in areas such as healthcare, where an accurate diagnosis is essential to ensure seamless user experience errors are not tolerated [6]. Research work on ChatGPT discuss its potential in different application domains. However, to our knowledge, no work covers its potential for translating sign language. The main contributions in our paper are as follows:

1. Depict a retrospective analysis of the evolution of ChatGPT architecture.
2. Design and develop experiments for translating various SL to spoken languages using examples from healthcare scenarios.
3. Present the requirements and challenges of ChatGPT for developing real-time applications, such as SLT.

The rest of the paper is organized as follows. Section II presents the related works. The evolution of GPT is discussed in Section III. Section IV outlines the architecture. Section V

Correspondence: leila@uaeu.ac.ae

delves into the requirements associated with ChatGPT. The challenges and proposed solutions are discussed in Section VI. Section VII presents the experiments, results, and numerical analysis. Lastly, Section VIII concludes the paper.

## II. RELATED WORK

Table 1 compares related works [8]–[11]. To our knowledge, we are the first to present a detailed architecture and the possible use of ChatGPT for SLT.

TABLE I. COMPARISON BETWEEN THE RELATED WORK AND THIS WORK.

| Work | Architecture | Application | Evolution | Limitations | Requirements | Challenges Domains | | |
|---|---|---|---|---|---|---|---|---|
| | | | | | | Energy Consumptio | Ethical | Privacy and Security |
| [8] | ✓ | - | ✓ | ✓ | ✓ | ✓ | ✓ | ✓ |
| [9] | ✗ | Healthcare | ✗ | ✓ | ✗ | ✗ | ✓ | ✗ |
| [10] | ✗ | Scientific Research | ✗ | ✓ | ✗ | ✗ | ✓ | ✗ |
| [11] | ✗ | Education | ✗ | ✓ | ✗ | ✗ | ✓ | ✓ |
| Our Work | ✓ | SLT | ✓ | ✓ | ✓ | ✓ | ✓ | ✓ |

## III. EVOLUTION OF GPT

GPT has evolved over the years, intending to provide accurate responses. Fig. 1 illustrates the timeline of GPT's evolution, Table 2 shows the training dataset types and sizes used to train/retrain the different GPT versions, and Table 3 compares GPT's architecture, functionality, and limitations.

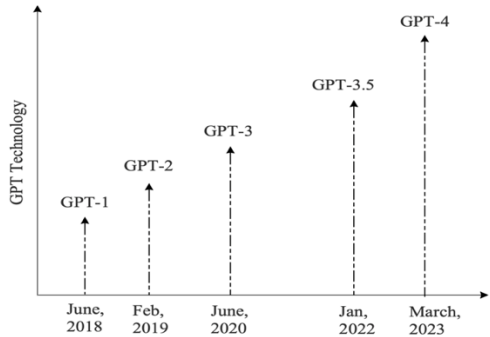

Fig. 1. GPT evolution over time.

## IV. ARCHITECTURE OVERVIEW

Fig. 2 shows use case study of English to ASL and vice-versa using the current version of ChatGPT architecture, which is composed of encoders, a decoder-only transformer [7], instruction fine-tuning [12], and reinforcement learning with human feedback [20] components.

### A. Encoders

There are two encoding procedures in ChatGPT architecture, with an embedding step in between, as presented in the following steps [13]:

A.1. The input sequence ("I broke my arms") is encoded as tokens using Byte Pair Encoding (BPE) [14] to create a vocabulary of sub-word tokens.
A.2. The tokens are embedded in a smaller dimensional space.
A.3. The embeddings are passed through Positional Encoding (PE) [7], as shown in Equations 1 and 2, where $pos$ denotes the current position, $i$ denotes the dimension index, $d$ denotes the dimension. For instance pos("I")=0 and PE("I")=(0, 1, 0, …,0, 1)..

$$PE(pos, 2i) = \sin\left(\frac{pos}{1000^{\frac{2i}{d_{model}}}}\right) \quad (1)$$

$$PE(pos, 2i + 1) = \cos\left(\frac{pos}{1000^{\frac{2i}{d_{model}}}}\right) \quad (2)$$

### B. Transformer

The transformer [7] processes NLP tasks such as language translation, as it handles long data sequences efficiently. It consists of a series of decoders [15], which generate the output sequence one token at a time, using the encoded representations as input. The transformer has the following layers:

B.1. The multi-head self-attention layers assess the importance of different words or phrases in each input. The first step in these layers is to attain the Query (Q), Keys (K), and Values (V) by passing the positional embeddings through three linear layers. Then create an Attention Head (AH) using Q and K to decide the attention needed for each position, as shown in Equation 3 [7]. The multi-head concatenates multiple self-attention heads, as shown in Equation 4.

TABLE II. TRAINING DATA TYPES USED IN ALL GPT VERSIONS.

| Dataset Type | GPT-1 | GPT-2 | GPT-3 | GPT-4 |
|---|---|---|---|---|
| Wikipedia | - | - | 11 GB | - |
| Books | 5 GB | - | 21 GB | - |
| Academic Journals | - | - | 101 GB | - |
| Reddit Links | - | 40 GB | 50 GB | - |
| Common Crawl | - | - | 570 GB | - |
| Total size | 5 GB | 40 GB | 753 GB | 20 TB |

TABLE III. COMPARISON BETWEEN GPT VERSIONS, SHOWING ADDITIONAL FEATURES REPORTED COMPARED TO THE PREVIOUS VERSION IN TERMS OF ARCHITECTURE, SERVICES, AND LIMITATIONS.

| | Version | GPT-1 [16] | GPT-2 [13] | GPT-3 [17] | GPT-3.5 [18] | GPT-4 [6] |
|---|---|---|---|---|---|---|
| Architecture | Number of Parameters | 117 million | 1.5 billion | 175 billion | Three variants: 1.3 billion (B), 6 B, and 175 B | 1 trillion |
| | Self-Attention | Yes (Masked and Multi-head) | | | | |
| | Vocabulary | 4,097 tokens | 50,257 tokens | 3,000B tokens | - | 32,768 tokens |
| | Hidden Layers | 768 | 1600 | 12,888 | - | - |
| | Decoder Layers | 12 | - | 96 | - | - |
| | Attention Head | 12 | 48 | 96 | - | - |
| | Feed Forward | 768 dimensions | 1600 dimensions | 12888 dimensions | - | - |
| | Window Size | 512 tokens | 1,024 tokens | 2,048 tokens | - | 8,195 tokens |
| | Optimizer | Adam | | | | |
| | Learning Rate | 2.5e-4 | - | $0.6 \times 10^{-4}$ | - | - |
| | Dropout | Rate of 0.1 | - | - | - | - |
| | Activation Function | GELU | - | - | - | - |
| | Epochs | 100 | - | - | - | - |
| | Batch Size | 64 | 512 | 3.2 million | - | - |
| Functionalities | | • Text Completion<br>• Language Translation<br>• Summarization<br>• Question Answering<br>• Sentiment analysis<br>• Prompt-based language generation | • Text-based conversational agent | • Language understanding for a specific domain<br>• Mathematical addition<br>• Article generation<br>• Vocabulary interpretation<br>• Code writing | - | • Pass bar exams<br>• Accept images as an input<br>• Browse the internet through plugins |
| Limitations | | • Limited reasoning<br>• I/O: text only<br>• Limited domain-specific knowledge<br>• Weak generalization<br>• Disconnected from internet | • Limited reasoning<br>• I/O: text only<br>• Limited domain-specific knowledge.<br>• Disconnected from internet | • Limited reasoning<br>• I/O: text only<br>• Responds to harmful instructions<br>• Capacity issues<br>• Disconnected from internet | • Limited reasoning<br>• I/O: text only<br>• Trained on Internet data before 2021<br>• Capacity issues<br>• Disconnected from internet | • Generates text only<br>• Trained on Internet data before 2021 |

$$AH = Sofmax(\frac{QK}{\sqrt{d}})V \qquad (3)$$

$$MultiHead = Concat(AH_1 \dots AH_i) \qquad (4)$$

B.2. The normalization layers are added to reduce the execution time [17].

B.3. The feed-forward layers (FFL) improve the representation of the relationships between different words or phrases. Equation 5 shows the FFL calculation, where $X_i$ denotes the input, $W_i$ denotes the weight, and $b_i$ denotes the bias of the i[th] layer.

$$FFL = ActivationFunction(X_i \cdot W_i + b_i) \qquad (5)$$

B.4. The output of the FFL is normalized and then fed into a linear layer to project the vector produced into a larger vector.

B.5. The Softmax layer that is in place to turn the vectors into probabilities. The highest probability is chosen, and the word associated with it is produced as the output for this time step.

*C. Instruction Fine-Tuning*

The instruction fine-tuning involves explicit instructions to fine-tune the model [12], which aims to improve the accuracy of the generated responses [9]. In this process, the output of the pre-trained model is combined with the decoder output [12].

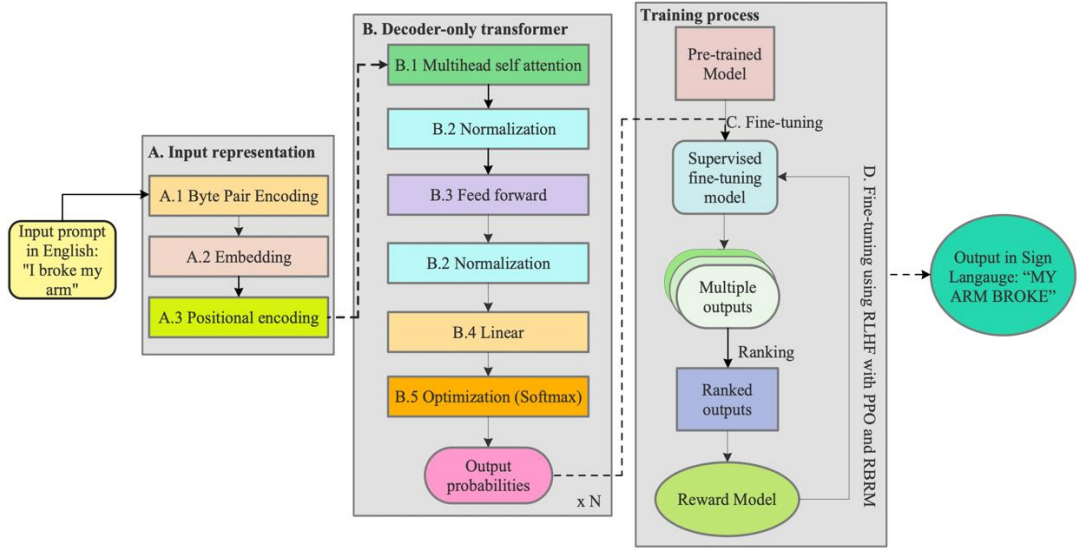

Fig. 2. GPT architecture for Sign Language Translation. RLHF refers to reinforcement learning with human feedback. PPO refers to the Proximal Policy Optimization.

### D. RLHF

RLHF trains ChatGPT by integrating human feedback into the learning process. RLHF is implemented using the PPO algorithm and the Rule-Based Reward Model (RBRM) [7]. The output of this step aligns better with human preferences and follows instructions more effectively.

D.1. The PPO algorithm is a model-free, on-policy algorithm. It learns from the environment through trial and error and updates its policy accordingly. PPO uses a clipped surrogate objective function, as shown in Equations 6 and 7, to encourage the policy to make significant updates when it is far from optimal yet limit the updates when it is close to optimal [18].

$$L^{CLIP}(\theta) = \hat{\mathbb{E}}_t[min(r_t(\theta)\hat{A}_t, clip(r_t(\theta), 1-\epsilon, 1+\epsilon)\hat{A}_t)] \quad (6)$$

$$r_t(\theta) = \frac{\pi_\theta(a_t|s_t)}{\pi_{\theta old}(a_t|s_t)} \quad (7)$$

where $\theta$ is the policy parameters, $r_t(\theta)$ is the ratio of the probability of the new policy to the old policy, $\hat{A}_t$ is the advantage function, and $\epsilon$ is a hyperparameter that defines the clip range.

D.2. RBRM is a model that defines the reward based on rules to provide a more informative and structured reward to aid in reaching the goal. This model defines the reward function as a set of Boolean conditional statements. The reward itself is a numerical value indicating how desirable the corresponding state or action is.

## V. ChatGpt Sign Langauge Applications Requirements

ChatGPT has the potential to be implemented in different fields, such as SLT. However, it must satisfy the following requirements:

- Accuracy: For language translation to be accurate, ChatGPT should be trained in sign languages grammar and content.
- Real-time [19]: Computing resources are needed for training and text generation.
- Data privacy and security: Measurements to handle sensitive information stored for processing.
- Adaptability: The input and output length should be adaptable based on the context.
- Support for multiple I/O formats: Needed for SLT applications which require a variety of I/O formats [20].

## VI. Challenges

The abovementioned requirements have different challenges, such as energy consumption, ethical consideration, and privacy and security.

### A. Energy Consumption

ChatGPT, of 1 trillion parameters, energy consumption is apprehensive. LLAMA, which has only 65B parameters, needs 2048 A100 GPUs and around 21 days to be trained [23]. Compared to ChatGPT (GPT-4). Consequently, ChatGPT needs about 323 days to be trained if it was trained on the same GPU. Such consumption can lead to pollution. A report published by the International Energy Agency (IEA) revealed that in 2021, global data centers consumed around 220-320 terawatt-hours of electrical energy, excluding the energy used for cryptocurrency mining. These numbers were estimated to triple by 2025 [21]. This report was published before the launch of ChatGPT.

### B. Ethical Consideration

Ethical problems can happen unintentionally during training. An example is the biased and discriminatory language in the training data. Such a challenge can be mitigated by considering fairness during the data collection and pre-processing.

### C. Privacy and Security

Security mechanisms should be put in place to ensure privacy while training on sensitive information and mitigate the security threats, such as data breaches, associated with ChatGPT and other language models [22].

## VII. EXPERIMENTS, RESULTS AND NUMERICAL ANALYSIS

### A. Experimental Setup and Experimentss

We run the translation experiments for different languages on the free version of ChatGPT (GPT-3.5) [23], as shown in Table 4. In these experiments, we consider ASL, AUSLAN, and BSL as one sign language because they all use English as text and share the same grammar and sentence structure. We ask ChatGPT to translate 5 medical-related statements which can be part of a conversation between a patient and a doctor or nurse, as shown in Table 6. To measure the performance of ChatGPT in one experiment, we count the number of prompt iterations needed by ChatGPT to achieve correct translations and calculate the accuracy as described below.

TABLE IV. EXPERIMENTS FOR TRANSLATING SIGN LANGUAGE TO SPOKEN LANGUAGE AND VICE VERSA

|  | AABSL Experiments | ArSL Experiments |
|---|---|---|
| English | 1. English-AABSL<br>2. AABSL-English | 3. English-ArSL<br>4. ArSL-English |
| Arabic | 5. Arabic-AABSL<br>6. AABSL-Arabic | 7. Arabic-ArSL<br>8. ArSL-Arabic |

AABSL: ASL/AUSLAN/BSL

#### 1) Prompt Iterations

For each translation experiment, i.e., English – AABSL, we initiate a prompt asking ChatGPT to act as a sign language interpreter and translate the sentences we provide. If the translation is incorrect, we correct the incorrect translation for each statement by initiating a new prompt iteration by providing the correct translation. Once ChatGPT translates a statement correctly, we stop and repeat the same steps for the next statement. We end the experiments after a maximum of 4 prompts iterations for each statement. Table 5 shows an example of the prompt iterations where the number of prompts is 1.

#### 2) Accuracy

We calculate the accuracy for each statement, as shown in Equation 8, and the total accuracy for each experiment, as shown in Equation 9. $SA_i$ is the ith statement accuracy, $PICT_i$ is the number of prompt iterations until the correct translation is achieved, EA is the experiment accuracy, and N is the total number of statements in an experiment.

TABLE V. EXAMPLE OF PROMPT ITERATIONS (ENGLISH TO AMERICAN SIGN LANGUAGE).

| Input prompt (0): | Act as a sign language interpreter. I will provide you with sentences to translate to sign language in text with the correct grammar and gloss. output the sign language text only without the steps to sign it or any other words such as "Gloss" or "Translation". |
|---|---|
| ChatGPT: | Sure! I can help you with that. Please provide me with the sentences you would like me to translate into sign language. |
| Input prompt (1): | I broke my arm |
| ChatGPT: | MY ARM BROKE |

$$SA_i = \begin{cases} \frac{1}{PICT_i} \times 100, 1 \leq PICT \leq 4, if\ successful \\ 0, \quad Otherwise \end{cases} \quad (8)$$

$$EA = \frac{\sum SA_i}{N} \quad (9)$$

### B. Numerical Results Analysis

We divide the results into two parts, 1) translation from spoken languages to sign languages, and 2) translation from sign languages to spoken languages. Fig. 3 and 4 present the number of prompt iterations until a correct translation is achieved and accuracy for each translation and experiment.

#### 1) Spoken Languages to Sign Languages

ChatGPT successfully translated the five statements from English to AABSL from the first iteration, which resulted in 100% accuracy, as shown in Fig. 3 and 4. This is due to the high number of books and online content for SL grammar and teaching material that was most probably included in GPT's training. On the other hand, instead of translating from English to ArSL, ChatGPT initially translated all statements to Arabic but managed to translate 2 statements correctly after prompt iterations. This resulted in 17% experiment accuracy. We assume that this is because of the lack of online ArSL content; to our knowledge, there are no complete ArSL dictionary and grammar resources yet.

Furthermore, ChatGPT could translate 3 statements from Arabic to AABSL after we interfered and corrected the translation. Although, it initially translated all statements to English rather than AABSL. The total accuracy for this experiment is 27%, as shown in Fig. 4. Lastly, ChatGPT failed to translate any statement correctly from Arabic to ArSL. After three prompt iterations, ChatGPT advised us to ask for a competent human interpreter to perform such tasks. Therefore, the accuracy for this translation is 0%. These results were due to the lack of ArSL online content and material. In addition, ChatGPT was trained better in English than Arabic.

TABLE VI. STATEMENTS USED IN THE EXPERIMENTS.

| # | English | AABSL | Arabic | ArSL |
|---|---|---|---|---|
| 1 | I broke my arm | MY ARM BROKE | كسرت ذراعي | ذراعي كسر |
| 2 | The doctor prescribed me medicine | DOCTOR GIVE-ME MEDICINE | الطبيب وصف لي الدواء | الطبيب وصف لي دواء |
| 3 | You need to do an X-Ray on the second floor | SECOND FLOOR X-RAY NEED DO | عليك أن تعمل أشعة سينية في الطابق الثاني | أشعة سينية الطابق الثاني يجب عمل |
| 4 | Should I book you a follow-up appointment next week? | NEXT WEEK FOLLOW-UP APPOINTMENT BOOK SHOULD I? | هل يجب أن أحجز لك موعد متابعة في الأسبوع القادم؟ | الأسبوع القادم موعد متابعة أحجز لك؟ |
| 5 | You should take this medicine twice a day after food | FOOD AFTER DAY-TWICE MEDICINE TAKE YOU SHOULD | يجب أن تتناول هذا الدواء مرتين في اليوم بعد تناول الطعام | بعد أكل مرتين يوم هذا الدواء خذ |

AABSL: American, Australian and British Sign Languages

*1) Sign Languages to Spoken Languages*

ChatGPT translated all statements from AABSL to English. However, after three iterations, it failed to translate any statement from AABSL to Arabic. Moreover, ChatGPT successfully translated all statements from ArSL to English and Arabic, despite translating 2 through passive rather than active voice. Based on Fig. 4, ChatGPT is more competent in translating from sign to spoken languages but performs better in English than Arabic.

VIII. CONCLUSION AND LESSONS LEARNED

In this paper, we present a retrospective analysis of the evolution of ChatGPT functionalities over time. We also investigate its architecture via a step-by-step approach to Sign Language Translation. Additionally, we perform experiments to translate from multiple sign languages to spoken languages and vice versa. The results show that ChatGPT can translate from Sign Languages to Spoken Languages. Additionally, it performs better when translating to English rather than Arabic.

*A. Lessons Learned*

- Enhancing GPT architecture is hindered by challenges and limitations, despite its vast potential.
- Arabic Sign Language translation requires significant improvements, contingent on the available corpus.
- ChatGPT can translate from spoken to sign languages, as text only, and vice versa. However, ChatGPT does not support visual sign language as input.

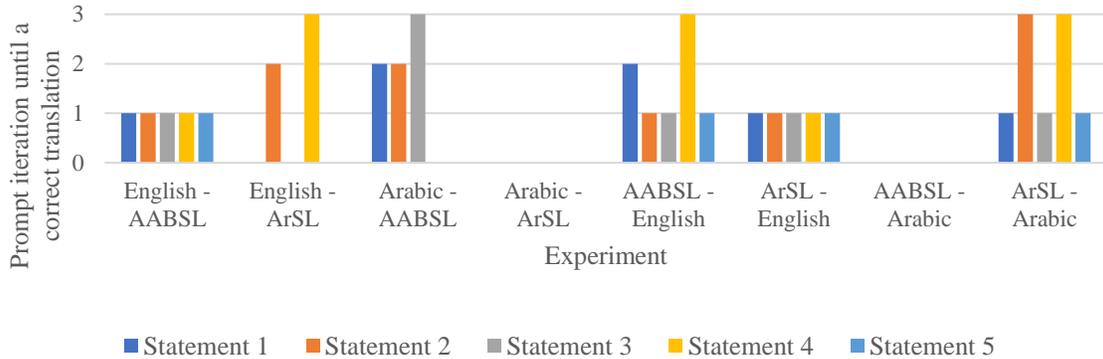

Fig. 3. Number of prompt iteration until a correct translation for each statement translation. AABS refers to ASL/AUSLAN/BSL.

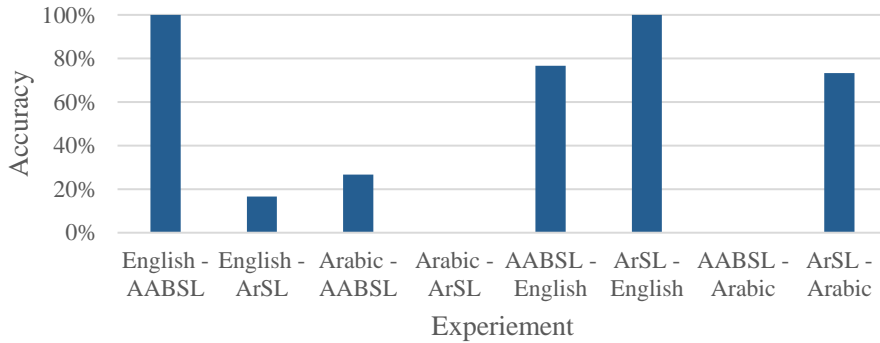

Fig. 4. Accuracy for each statement translation and experiment. AABSL refers to ASL/AUSLAN/BSL.

ACKNOWLEDGMENT

This work was supported by the National Water and Energy Center of the United Arab Emirates University under grant 12R056.

REFERENCES

[1] World Health Organization, "Deafness and hearing loss," *World Health Organization*, 2023. https://www.who.int/news-room/fact-sheets/detail/deafness-and-hearing-loss (accessed May 29, 2023).

[2] Zippia, "Interpreter Demographics and Statistics in the US," 2022. https://www.zippia.com/interpreter-jobs/demographics/ (accessed May 29, 2023).

[3] National Institute on Deafness and Other Communication Disorders, "Quick Statistics About Hearing," 2021. https://www.nidcd.nih.gov/health/statistics/quick-statistics-hearing (accessed May 29, 2023).

[4] G. Z. de Castro, R. R. Guerra, and F. G. Guimarães, "Automatic translation of sign language with multi-stream 3D CNN and generation of artificial depth maps," *Expert Syst Appl*, vol. 215, p. 119394, Apr. 2023.

[5] N. Shahin and M. Watfa, "Deaf and hard of hearing in the United Arab Emirates interacting with Alexa, an intelligent personal assistant," *Technol Disabil*, vol. 32, no. 4, pp. 255–269, Nov. 2020.

[6] OpenAI, "GPT-4 Technical Report," 2023.

[7] A. Vaswani, N. Shazeer, N. Parmar, and J. Uszkoreit, "Attention is All you Need," *Advances in Neural Information Processing Systems*, 2017.

[8] T. Wu *et al.*, "A Brief Overview of ChatGPT: The History, Status Quo and Potential Future Development," *IEEE/CAA Journal of Automatica Sinica*, vol. 10, no. 5, pp. 1122–1136, May 2023.

[9] M. Cascella, J. Montomoli, V. Bellini, and E. Bignami, "Evaluating the Feasibility of ChatGPT in Healthcare: An Analysis of Multiple Clinical and Research Scenarios," *J Med Syst*, vol. 47, no. 1, p. 33, Mar. 2023.

[10] C. Macdonald, D. Adeloye, A. Sheikh, and I. Rudan, "Can ChatGPT draft a research article? An example of population-level vaccine effectiveness analysis," *J Glob Health*, vol. 13, p. 01003, Feb. 2023.

[11] A. Tlili *et al.*, "What if the devil is my guardian angel: ChatGPT as a case study of using chatbots in education," *Smart Learning Environments*, vol. 10, no. 1, p. 15, Feb. 2023.

[12] H. W. Chung *et al.*, "Scaling Instruction-Finetuned Language Models," Oct. 2022.

[13] A. Radford, J. Wu, R. Child, and D. Luan, "Language models are unsupervised multitask learners," *OpenAI Blog*, vol. 1, no. 9, 2019.

[14] P. Gage, "A new algorithm for data compression," *C Users Journal*, vol. 12, no. 2, pp. 23–38, 1994.

[15] R. Child, S. Gray, A. Radford, and I. Sutskever, "Generating Long Sequences with Sparse Transformers," Apr. 2019.

[16] A. Radford, K. Narasimhan, T. Salimans, and I. Sutskever, "Improving Language Understanding by Generative Pre-Training," 2018.

[17] T. Brown, B. Mann, N. Ryder, M. Subbiah, and J. Kaplan, "Language Models are Few-Shot Learners," in *34th Conference on Neural Information Processing Systems*, 2020.

[18] T. Hagendorff, S. Fabi, and M. Kosinski, "Machine intuition: Uncovering human-like intuitive decision-making in GPT-3.5," *ArXiv*, 2022.

[19] L. Ismail, H. Materwala, A. P. Karduck, and A. Adem, "Requirements of Health Data Management Systems for Biomedical Care and Research: Scoping Review," *J Med Internet Res*, vol. 22, no. 7, 2020.

[20] L. Ismail and M. D. Waseem, "Towards a Deep Learning Pain-Level Detection Deployment at UAE for Patient-Centric-Pain Management and Diagnosis Support: Framework and Performance Evaluation," *Procedia Comput Sci*, vol. 220, pp. 339–347, 2023.

[21] International Energy Agency, "Data Centres and Data Transmission Networks," 2022. Accessed: May 29, 2023. [Online]. Available: https://www.iea.org/reports/data-centres-and-data-transmission-networks

[22] S. Addington, "ChatGPT: Cyber Security Threats and Countermeasures," *SSRN Electronic Journal*, 2023.

[23] OpenAI, "ChatGPT," 2022. https://chat.openai.com (accessed May 31, 2023).